REVIEW ARTICLE

Farm security · Rural crime · Livestock predation · Camera-trap AI · Evidence review

# The Case for AI-Enabled Gate Monitoring as Farm Security in the UK: A PRISMA-Guided Review of the Evidence on Rural Crime, Livestock Predation and Farm Access

Dr Steve Symes[1], Prof Paul Fergus[1], Mr Mark Carr[2], Mr Omar McIver[3] and Mr Arwyn Jones[4]

[1]School of Computer Science and Mathematics, Liverpool John Moores University, Liverpool, UK. [2]Image Surveys Limited, Southport, UK. [3]Mode3 Cloud, Katy, Texas, USA. [4]Labour Force Europe Ltd, Corwen, North Wales.

Correspondence: S.W.Symes@ljmu.ac.uk , P.Fergus@ljmu.ac.uk , Mark@imagesurveys.com , Omar.McIver@mode3.ai , arwyn@labourforceeuropeltd.com

## ABSTRACT

Farm security in the UK rests on thinner evidence than its cost suggests. We ran a PRISMA guided review: twelve searches of Web of Science and Scopus on 2nd -4th November 2026 returned 1,183 records (761 unique), and grey-literature and hand-searches added 35 assessed sources. We synthesised 96 database records and 24 grey sources narratively. UK peer-reviewed evidence is small (16 records), and the main UK survey of crime and farmer mental health has 80 responses. Cost estimates rest on insurer claims: NFU Mutual put UK rural crime at £41.5 million in 2025, with livestock theft up almost 30% to about £4.5 million and dog attacks on livestock costing £1.95 million. The Home Office states that agricultural crime cannot be separately identified in its central data. We found no independent evaluation of cameras, sensors, trackers or marking; light and sound deterrents lost effect within one to five months in a systematic review of anti-predator interventions; and no study examined how often farm gates are left open or what follows. Camera-trap AI papers report accuracy on ecological datasets, none at a farm gate in the records we found, and in one comparison machine learning detected 41% of wildlife against 81% for analysts. We argue, as an inference from these gaps and not as a demonstrated effect, that continuous gate-level AI-mediated monitoring, as implemented in Fieldmark AI, would be highly beneficial for UK farm security. It would create the farm-level, time-stamped record that no official source holds, verify events that periodic human checks miss, resolve whether foxes, dogs or people were responsible, and give every other security measure something to be tested against. No study has yet shown that it reduces loss, and we set out the evaluation needed to show it.

## KEY POINTS

- The evidence on farm security is thin where it matters most to a UK farmer: only 16 of 96 database records are UK peer-reviewed studies, and the principal UK survey linking crime to farmer mental health has 80 responses.
- Cost figures come from insurer claims (£41.5 million in 2025), because police-recorded data in England and Wales cannot identify farm crime.
- No independent evaluation of cameras, sensors, trackers or marking was found, and light and sound deterrents lost effect within one to five months in a systematic review. Physical barriers and management are better supported.
- Camera-trap AI has been validated on ecological data, and we found no farm-gate evaluation. No

study examined how often farm gates are left open or what follows.

- The authors argue that continuous, gate-level AI monitoring such as Fieldmark AI would be highly beneficial because it fills the largest of these gaps. This is an inference, not a demonstrated reduction in loss, and Box 1 sets out how to test it.

# 1 Introduction

Farm security in the United Kingdom (UK) is usually discussed in terms of cost. NFU Mutual, the largest rural insurer, estimated the cost of UK rural crime at £41.5 million in 2025, with livestock theft up almost 30% to about £4.5 million [1]. It separately estimated £1.95 million of farm animals severely injured or killed by dogs in 2025, up 10% [2]. Those are claims-based insurer figures. The Home Office has stated that offences involving agricultural crime "cannot be separately identified" in the data it holds centrally [3], and that there is no distinct offence category that captures rural crime [4]. The costs are therefore reported with confidence, while the events behind them are not systematically recorded.

Farm security is also wider than crime. A farm is a long, open boundary with many gates. Theft of livestock and machinery sits alongside predation by foxes and other species, attacks by dogs, trespass, and the access that walkers and vehicles have through gates and across fields. Each of these is a question of what happened, when, and who or what was responsible, and each is hard to answer from a periodic visit to the gate.

This review asks what the evidence supports about farm security, and whether it substantiates continuous, AI-mediated monitoring of farm gates as part of it. It is the second of two reviews by the authors. The first examined the economic, regulatory and technological case for integrated, AI-enabled land management in UK agriculture [5]. A companion case study reports a 14-camera gate-monitoring deployment at a Welsh hill farm on the Fieldmark AI platform [6]. This review supplies the evidence context for that deployment. The deployment is not itself evidence in the review.

We pose five questions.

- **RQ1.** What is known about the scale, nature and cost of theft, predation, dog attacks and unauthorised access on UK farms?
- **RQ2.** How reliable are the data: how is farm crime recorded and reported, and how far do farmers trust the system?
- **RQ3.** What evidence exists on the effectiveness of prevention and detection measures, including cameras and sensors?
- **RQ4.** What evidence exists that camera-trap-style AI can detect the targets that matter to farm security (people, vehicles, predators, dogs), and with what limits?
- **RQ5.** Given the answers, which evidence gaps would continuous gate-level monitoring address, and what would be needed to show that it works?

Section 2 sets out the methods. Section 3 reports the results by domain, ending with the contrary and null findings. Section 4 discusses what the evidence establishes, makes the case for continuous gate-level monitoring and states what would test it. Sections 5 and 6 give the limitations and conclusions.

# 2 Methods

## 2.1 Design and reporting

We conducted a literature review guided by the PRISMA statement [7] and reported the searches against PRISMA-S [8]. The searches are systematic and the review is structured, but it departs from a full systematic review in four ways, stated here and in Section 5. Screening and data extraction were carried out by a large language model under the authors' direction, and the authors checked its decisions. Data were extracted from titles, abstracts and texts, with full texts read for key records. The twelve database searches were all run between 2$^{nd}$ – 4$^{th}$ November 2026 and were completed before screening, extraction or writing began. The grey-literature and hand-searches followed the first screening pass and were aimed at gaps it exposed. A mapping of the PRISMA items to this paper is in Appendix B.

## 2.2 Eligibility criteria

We included studies, reports and official or professional documents that gave data, analysis or evaluation on one of seven domains: **D1** theft of livestock and equipment; **D2** predators and dogs; **D3** gates, public access and straying; **D4** prevention and detection other than AI; **D5** camera-trap AI; **D6** under-reporting, recording, trust and data quality; and **D7** costs and impact on farmers. Farmers, farms or agricultural premises had to be the population or setting. Records had to be published between 2000 and 2026 in English. Peer-reviewed articles, reviews and book chapters were eligible, as were conference papers for the technology questions. Grey sources were eligible if they came from government, police, regulators, parliaments, farming bodies, insurers or universities.

We excluded ecology and conservation studies with no farm or livestock link, large-carnivore studies that measured no livestock outcome, crop-pest and yield studies with no security aim, opinion pieces with no data (unless they supplied a unique UK fact), duplicates, and non-English or pre-2000 records. Reviews of predation interventions in large carnivores were kept as evidence-quality framing for the UK-relevant evidence.

## 2.3 Information sources and search

We searched the Web of Science Core Collection (all editions) and Scopus (advanced search) with six queries each, one per topic area (Table 1; full strings in Appendix A), limited to 2000–2026 and English. Query 2 excluded large-carnivore terms from titles. Web of Science hit counts before refinement were not recorded, and all Scopus hits were exported. The grey-literature arm comprised 29 web searches of official, police, regulatory, parliamentary, insurer and farming-body sources. It retrieved and assessed 35 sources, including two articles from the *International Journal of Rural Criminology* that neither database indexes [9,10]. We did not search Google Scholar, CAB Abstracts, Criminal Justice Abstracts or ProQuest (Section 5).

**Table 1 | Database searches run on 4 November 2026.** Records exported per query from Web of Science (WoS) and Scopus; 'held' means already in the Web of Science set or an earlier Scopus query.

| No. | Topic | WoS exported | Scopus exported | Scopus held | Scopus new |
|---|---|---|---|---|---|
| 1 | Theft of livestock and farm equipment | 67 | 111 | 60 | 51 |
| 2 | Predation and livestock worrying | 415 | 364 | 292 | 72 |
| 3 | Gates, straying and public access | 17 | 31 | 15 | 16 |
| 4 | Crime prevention and detection technology | 25 | 66 | 21 | 45 |
| 5 | Camera-trap AI | 31 | 43 | 23 | 20 |
| 6 | Under-reporting and data quality | 2 | 11 | 8 | 3 |

| No. | Topic | WoS exported | Scopus exported | Scopus held | Scopus new |
|---|---|---|---|---|---|
| | **Total** | 557 | 626 | 419 | 207 |

## 2.4 Selection process

Records were de-duplicated by DOI or normalised title. A single reviewer, a large language model (Claude, Anthropic) working from the exported titles, abstracts and texts under the authors' direction, assigned each unique record Include, Maybe or Exclude with a reason and an evidence tier. The 92 Maybe records were then re-read against a gap rule. A Maybe was promoted if it was among the few sources on a domain that the first-pass evidence map showed to be thin (camera-based detection of farm targets, the durability or independent testing of deterrents, the quality of predator-intervention evidence, public access, or UK methods), and excluded if it duplicated stronger evidence or was off topic. The rule was set after the first pass and is therefore post hoc.

## 2.5 Data extraction and evidence tiers

For each included record we extracted setting, design, sample, evidence type, primary and other domains, the key finding, limitations, and how the record is used (directly for the UK, transferable from elsewhere, or context). Extraction was from the title, abstract and text, and where an abstract was silent the field says so. Full texts were read for all UK records with relevance as per and for many grey sources, using an automated reader (figures from those sources were checked by the authors against the originals). Two book chapters had no abstract and could not be extracted [14,15]. Fourteen overlaps or possible overlaps between records (for example two papers reporting the same Irish survey) were flagged and the samples are counted once in quantitative statements.

Each record was placed in an evidence tier by setting: **Tier 1**, UK; **Tier 2**, other high-income countries; **Tier 3**, other countries; and **Tech**, technology development papers whose setting is not a farm deployment. The tiers describe how directly a record applies to a UK farm and are not a ranking of quality.

## 2.6 Synthesis and appraisal

We synthesised the evidence narratively by domain and mapped it by domain and tier. We did not pool results. We did not apply a formal risk-of-bias tool or grade certainty. Instead each record carries its evidence type and limitations, and we describe the strength of each body of evidence in words. Findings that cut against the case for technology, and findings that are null or weak, are reported prominently (Section 3.9 and Table 4). The same search, screening and reporting rules were applied to evidence that supports and evidence that cuts against continuous monitoring.

## 2.7 Language-model screening and extraction

A large language model screened all 761 records and extracted the data, working under the authors' direction from records the authors exported from Web of Science and Scopus. The authors checked the model's screening and extraction decisions against the source records for accuracy.

# 3 Results

## 3.1 Study selection and characteristics

The searches returned 1,183 records (557 from Web of Science, 626 from Scopus). After 422 duplicates were removed, 761 unique records were screened on title and abstract: 588 were excluded, 81 included and 92 held as Maybe. On re-reading against the gap rule, 15 Maybes were promoted, which gave 96 database records. Of 35 grey sources assessed, 24 were included, 10 were kept as context only and one was not relied on. In all, 120 sources entered the synthesis (Fig. 1).

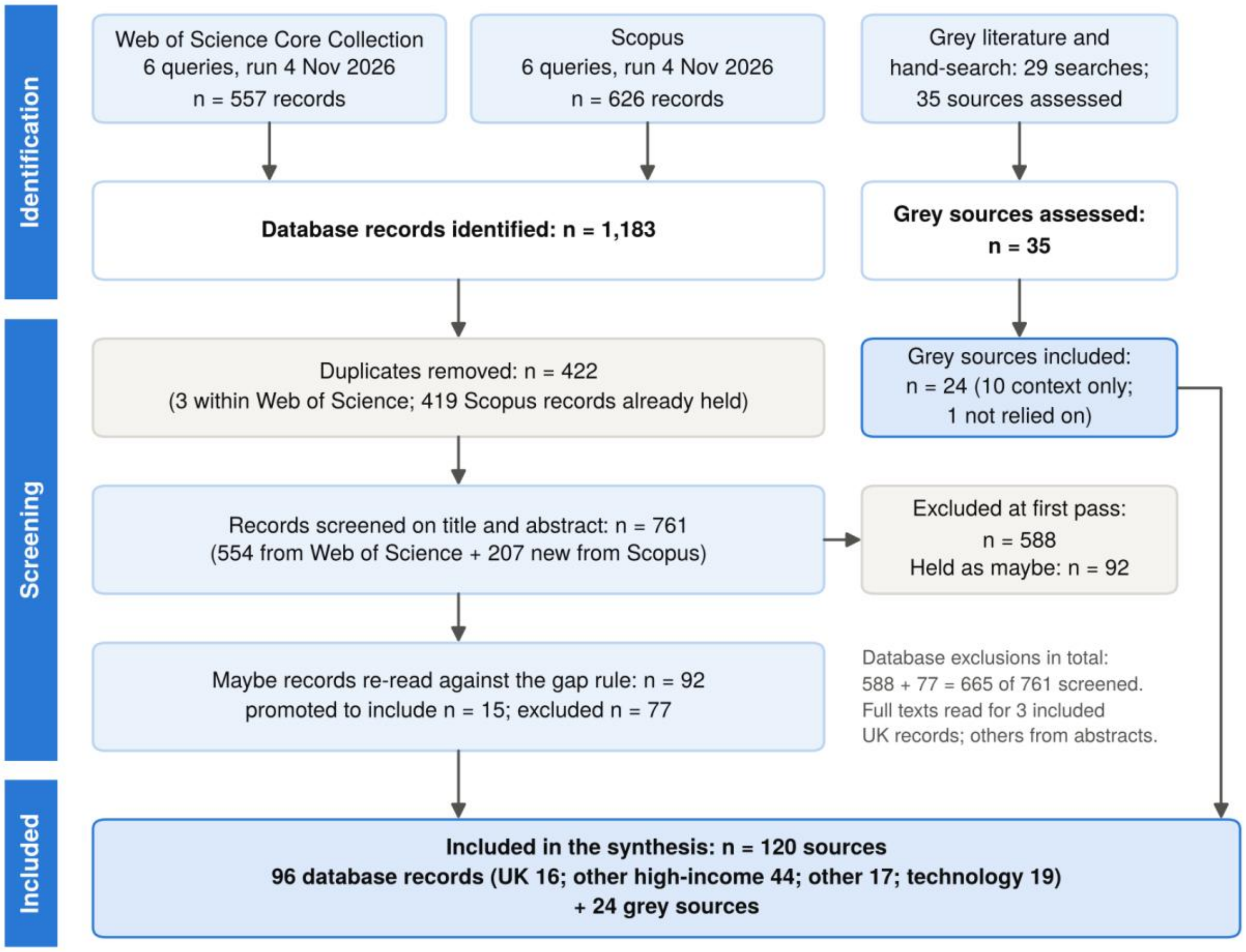

**Fig. 1 | PRISMA flow diagram.** Database arm (left) and grey-literature arm (right). The 15 promoted records were Maybes resolved by the gap rule in Section 2.4.

By evidence tier the 96 database records were 16 UK (Tier 1), 44 other high-income (Tier 2), 17 other non-UK (Tier 3) and 19 technology papers. Volume has grown fast: 65 of 96 records (68%) date from 2020 or later, and only seven appeared between 2005 and 2014 (Fig. 2). UK peer-reviewed evidence is small and recent, with nine of its 16 records published since 2020.

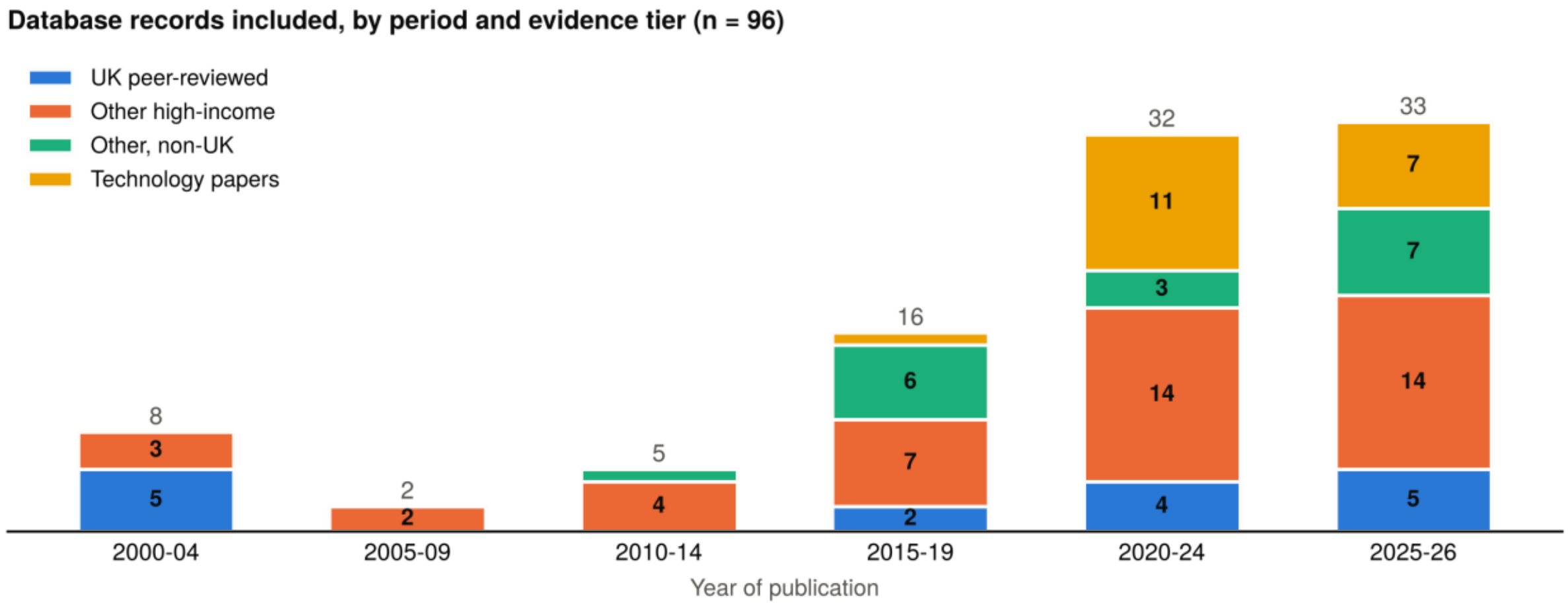

**Fig. 2 | Database records included, by period of publication and evidence tier.** Segments without a number are single records. Technology papers are those that develop or test detection systems rather than describe a farm deployment.

The evidence map (Fig. 3) shows where the weight lies. Predators and dogs hold the most database records (28), and theft the next most (18). Gates, public access and straying hold one (a study of livestock-related fatalities), and camera-trap AI holds 12, all of them technology papers or reviews. No database record evaluates a camera-based system at a farm gate. UK peer-reviewed evidence is concentrated in predators (seven records) and is absent for prevention and detection and for camera-trap AI.

| | | UK peer-reviewed | Other high-income | Other, non-UK | Technology papers | UK grey sources |
|---|---|---|---|---|---|---|
| D1 | Theft of livestock and equipment | 2 | 9 | 7 | 0 | 5 |
| D2 | Predators and dogs | 7 | 19 | 2 | 0 | 5 |
| D3 | Gates, public access, straying | 1 | 0 | 0 | 0 | 2 |
| D4 | Prevention and detection (non-AI) | 0 | 4 | 4 | 8 | 4 |
| D5 | Camera-trap AI | 0 | 1 | 0 | 11 | 0 |
| D6 | Under-reporting, trust, data quality | 3 | 8 | 2 | 0 | 6 |
| D7 | Costs and impact on farmers | 3 | 3 | 2 | 0 | 2 |

Counts are database records by primary domain (96) and grey sources marked Include (24). Hatched cells are empty.

**Fig. 3 | Evidence map.** Database records by primary domain and tier, and grey sources marked Include. Hatched cells are empty. D3 (gates, public access, straying) and D5 (camera-trap AI) are the thinnest domains for farm security.

## 3.2 Scale, nature and cost of farm crime (D1)

**The insurer series.** The most-used UK cost estimate is NFU Mutual's claims-based series, which fell from £52.8 million in 2023 to £44.1 million in 2024 and £41.5 million in 2025 [1,16]. Beneath the falling total, categories moved differently (Fig. 4). In 2025 agricultural vehicle theft cost £8.3 million (up 18%), quad bike and ATV theft £3.5 million (up 31%) and livestock theft about £4.5 million (up nearly 30%), described as highly organised, while GPS-unit theft fell 80% to about £250,000 [1]. Regionally the 2025 figures were £2.2 million in Wales (from £2.8 million), £1.4 million in Scotland (from £1.2 million) and £2.2 million in Northern Ireland (from £1.8 million) [1]. The 2024 report gave livestock theft as £3.4 million in its text and £3.3 million in its infographic [16].

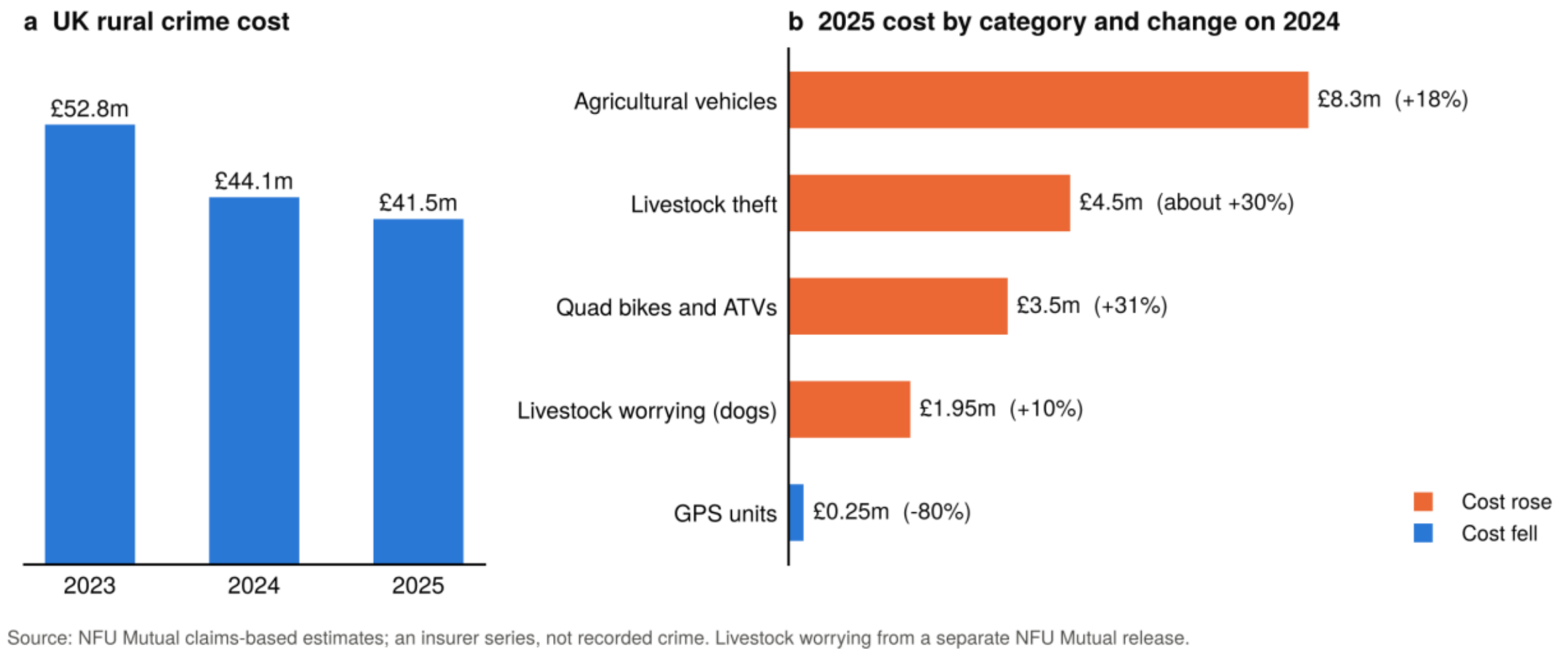

**Fig. 4 | UK rural crime cost. a**, Total estimated cost, 2023–2025 [1,16]. **b**, 2025 cost by category with the change on 2024 [1,2]. All figures are claims-based insurer estimates, not recorded crime.

**What the insurer series cannot tell us.** It covers insured losses only, the method of scaling from NFU Mutual claims is not given on the summary page, and an insurer has a commercial interest in the rural-risk narrative [1]. A Harper Adams commentary argues that the true cost is many times higher, because NFU Mutual insures about three-quarters of UK farmers and research it cites suggests only about 40% of crimes are reported to insurers, and that the 6% fall may be cyclical. Her extrapolations are the author's own assumptions, not data [17]. The two readings differ by an order of magnitude and the sources cannot arbitrate between them.

**Official and survey data.** The Home Office Commercial Victimisation Survey 2017 estimated that 27% of agriculture, forestry and fishing premises in England and Wales were victims of crime (30% in 2013), about 113,000 crimes, of which vandalism was about 37,000, theft 28,000 and burglary 18,000. The sector includes forestry and fishing, and farming is the main activity for 780 of 1,019 sampled premises [18]. An earlier press report of the 2014 survey gave about a quarter of premises affected [19]. In Northern Ireland, police recorded 205 agricultural crimes in the year to March 2026, counting only burglary, robbery and theft with an agricultural link [20]. Self-selecting national surveys give larger numbers with larger uncertainty: the 2015 National Rural Crime Survey (over 17,000 respondents) put the average cost of crime to a rural business victim at £4,100 [21], and the 2018 survey (20,252 responses) put the average cost per offence at £4,800, up 13% [22]. Table 2 collects the main UK figures and their limits.

**Table 2 | Principal UK figures on farm crime, with their limits.** Sources are cited by number; the figures differ in what, where and when they measure and are not additive.

| Measure | Figure | Source type | Limits |
|---|---|---|---|
| Cost of UK rural crime, 2025 | £41.5m, down 6% on 2024 (£44.1m); £52.8m in 2023 [1,16] | Insurer claims | Insured losses only; insurer has a commercial interest |
| Livestock theft cost, 2025 | About £4.5m, up nearly 30% [1] | Insurer claims | Summary page; full report not retrieved |
| Livestock worrying cost, 2025 | £1.95m, up 10% [2] | Insurer claims and owner survey | Sample sizes not given |
| Premises victimised, agriculture, forestry and fishing, E&W, 2017 | 27% (30% in 2013); about 113,000 crimes [18] | Home Office survey | Includes forestry and fishing; small sector sample |
| Agricultural crimes recorded, Northern Ireland, year to March 2026 | 205 (201 the year before) [20] | Police flag | Theft, burglary and robbery only |
| Dog attack or worrying incidents recorded, Scotland, year to March 2025 | 216, against 350 the year before [23] | Police Scotland via trade body | Recorded incidents only |
| Sheep farmers reporting a dog attack on their flock, 2024 | 87% [24] | NSA survey quoted by Defra | Survey not retrieved |
| Welsh farmers who had experienced crime | About half; 91% of victims reported it [11] | Survey, n = 96 | Recruited mainly at a show; descriptive only |
| Welsh farms victimised in past year | 22%; machinery theft 65% of crimes [25] | Survey, over 150 | News release only; report not located |
| British farmers who were victims | 72 of 80 (90%) [12] | Self-selected online survey | 5.1% response; prevalence cannot be inferred |
| Rural business victims, E&W | Average cost £4,100 (2015) [21]; £4,800 per offence (2018) [22] | Self-selecting surveys | Not farm-specific; may overstate |

**UK primary evidence is thin.** Two database records are UK peer-reviewed studies of theft: a horizon scan of emerging rural crime in England [26], which argues that English rural crime research is thin compared with that in the USA and Australia, and a multi-modal survey [27] (a chapter whose abstract gives no figures) covering victimisation, under-reporting and trust. A 2017 scoping survey, found by hand-search, had 71 complete responses from a snowball sample [10], and a methods paper sets out the advantages and disadvantages of online surveys of British farmers [28]. In Wales a farm crime survey found that machinery made up about half of crime and livestock 25% [11]. The practical consequence is that a UK farmer or policy maker cannot read a livestock-theft rate from the literature. NFU Cymru says livestock theft is getting worse in Wales, but gives no figures [29].

**Mechanisms that transfer.** Other countries supply mechanisms rather than rates. Australian surveys found that burglary and theft of fuel and machinery were inversely related to how visible farm buildings were from the farmhouse, and that stock theft rose with distance from town and with hilly terrain [30,31]; US work found that opportunity theory only partly explains victimisation [32]. A Pakistani case series of 53 farmers whose cattle were stolen found theft mostly at night, with repeat victimisation, and all respondents willing to invest in RFID tagging [33]. In South Africa, provincial police records for 2018–2024 show 30,816 stolen cattle in the Eastern Cape (R399 million), though species and recovery were weak predictors [34], and theft was significantly more likely in winter [35]. Surveys in Ireland, Slovenia, the USA, Australia, Nigeria and New Zealand report high victimisation (just over half of respondents in Ireland, 66% in north-west Nigeria) and, in several, under-reporting [36–41]. A Caribbean review finds farm theft hard to prove and lightly penalised [42], Swedish media archives show suspects are not one group [43], South African communities and farmers explain livestock theft differently [44], and South African farmers spend heavily on security with no single solution [45]. Equipment theft has its own policy track: the Equipment Theft (Prevention) Act will allow regulations on marking, registration and removable GPS units for new quad bikes and ATVs, of which 900 to 1,200 are stolen each year in England and Wales, but it has no effect until regulations are commenced [46].

## 3.3 Predators and dogs (D2)

**Foxes: small share, hard attribution.** The UK evidence on foxes suggests a smaller effect than farmer perception. On two Scottish hill farms lamb mortality was 9.3% to 10.2%, but confirmed fox predation was at least 0.2% to 0.6% of lambs and 53 lambs went missing, which cannot be attributed [47]. In a British survey, 59% of sheep farmers had lost at least one lamb to a fox at their last lambing, with reported losses of 0.0008 to 0.26 lambs per ewe [48]. Losses were perceived rather than confirmed, and the effect of fox control could not be determined because farmers react to losses [48]. A Scottish study combining farmer interviews, post-mortems and DNA found predation in 14 of 29 lambs (48%) with fox DNA on 87% of swabbed lambs, dog DNA on 31% and badger DNA on 5%, and no direct evidence that badgers kill lambs [49] but Badgers killing lambs has been witnessed by Arwyn Jones (an author on this paper and farmer in Wales). For free-range poultry, mean mortality was under 2% and fox predation was not linked to fox density, housing or fencing type [50].

**Dogs: rising cost, patchy records.** NFU Mutual estimated £1.95 million of livestock losses to dogs in 2025 (up 10%), and a dog-owner survey it cites found 57% let dogs off the lead in the countryside and one in ten admit no recall [2]. A National Sheep Association survey quoted by Defra found 87% of sheep farmers reported a dog attack on their flock in 2024 [24]. In Scotland, Police Scotland recorded 216 incidents in the year to March 2025 against 350 the year before [23]; an earlier briefing records 285 incidents in 2018/19 and 265 in 2019/20, and quotes Scottish Government research in which about 14% of nearly 2,000 sheep farmers had experienced worrying in the previous 12 months and around a third of attacks were reported [51]. A review of news coverage concluded that sheep worrying appears to be increasing and that prevention methods have not been evaluated [52]. The legal position changed on 18 March 2026, when the Dogs (Protection of Livestock) (Amendment) Act 2025 came into force in England and Wales with unlimited fines (previously £1,000), police powers to seize dogs and take DNA samples, and extended coverage of roads and paths [24,53]. The Welsh strategy describes a DNA project comparing attacked animals with suspected dogs [54]. Elsewhere, Sweden found that dogs accounted for 3.8% of inspected predator attacks and were usually loose or unsupervised [55], a survey of agriculture-ministry professionals in Chile found free-ranging dogs attacked 25% of small farms in a year [56], and in Uruguay over 85% of 91 sheep farmers reported predator attacks and distress was higher among farmers whose animals were attacked [57,58].

**Cameras reveal losses that records miss.** Two Australian field studies used cameras as the measuring instrument. At an outdoor piggery, about 20% of piglets were recorded as missing, and cameras caught foxes taking live piglets soon after staff left [59]; around guardian-dog paddocks, foxes appeared on camera on 17 of 23 nights and the dogs mostly guarded the poultry shelters, not the paddock [60]. Neither study set out to evaluate a security product, but both show that continuous observation can answer questions that periodic records cannot.

## 3.4 Gates, public access and straying (D3)

This is the thinnest domain. One database record has public access as its main topic: an analysis of Health and Safety Executive fatal-injury reports found 78 livestock-related deaths in Great Britain from 2010 to 2023, 97% involving cattle. Every member of the public killed was in a field, and 84.2% had a dog with them; the public were 250 times more likely than farmers to have a dog present [13]. It counts fatalities only, has no exposure denominator, and mentions gates only as handling infrastructure. HSE guidance says gates at entry points should be as high and strong as the perimeter fence, with a securing device that children and animals cannot release, and that fences, gates and stiles should be checked at least once a day [61]. It adds that many incidents go unreported [61]. Press reporting of HSE guidance and HSE statistics give the scale of public deaths in agriculture-related incidents (about one to two a year from cattle on rights of way; four in 2024/25 provisional, all causes) [62,63]. The Countryside Code asks the public to leave gates as they find them and to keep dogs under control, but the edition read has no livestock guidance [64]. Trespass appears as a secondary finding: 66.7% of 72 victims in one British survey reported it [12], and in Swedish media archives vandalism and trespass were common at mink farms [43].

No study found measured how often gates are left open, what follows when they are, or whether any intervention changes it. Queries 3 in both databases returned 48 records and none was about gates, straying or access in this sense.

## 3.5 Prevention and detection other than AI (D4)

**Evidence of effect is weak or negative.** The largest survey-based test of security practices found no association with lower victimisation: in a mail survey of 1,100 New South Wales farmers, crime was highly situational and security practices were not linked to lower victimisation [31]. A US programme evaluation suggested that marking equipment and livestock, surveillance, locked storage and prosecution may reduce victimisation, with the authors cautioning that design limits the finding [65]. Everything else on theft prevention is adoption, perception or a prototype. Irish farmers support prevention but do not widely practise it [36], some cannot afford it or lack the time [66], and Slovenian farmers rely on their own situational prevention [37]. In Wales, over 80% of respondents took precautions, but the study did not test whether they were linked to victimisation [11]. A union poll of nearly 2,000 farmers reported that 35% had installed CCTV, more than half had upgraded building security, and 31% had spent at least £1,000 on crime prevention over five years [67]. A survey of police forces (33 of 38 contacted responded) found technology including CCTV widely used and just over half of respondents rating methods effective, based on opinion, not outcomes [9].

**No independent evaluation of cameras, sensors, trackers or marking.** Insurer reports recommend CCTV, lighting, alarms, trackers, immobilisers and marking as measures "proven to deter determined criminals" but give no data on effect, and one case study in the same report shows thieves knowing where the CCTV was and footage capturing little [16]. Police Scotland recovered £893,000 of equipment in 2019 and a press report credits trackers, but there is no tracked-versus-untracked comparison, and an insurer discount bundles marking, immobiliser and tracker so the effect of the tracker cannot be isolated [68]. The Home Office call for evidence records respondents saying that etching and microdot marks can be removed and that no provider yet meets the relevant standard [46]. A trade claim of a 0.06% theft rate for marked machinery has no source, method or period, and we did not rely on it [69]. Other police and policy sources are context: a national rural crime unit funded in part externally and by an insurer [70], a national vehicle-crime intelligence service whose published cases include no agricultural theft statistics [71], a national policing strategy known only through a summary, which says technology will support intelligence-led policing [72], and a Welsh strategy that sets no numeric targets and describes no system for recording farm crime beyond "centralised intelligence" [54].

**Prototypes report high lab accuracy and no theft outcomes.** A Bangladeshi multi-sensor system detected 100% of intrusions with its laser sensor, up to 90.5% with passive infrared sensors at night, and an average efficiency of 92.5% by day and 94.38% at night, tested in its developers' own conditions with no theft outcome [73]. A related system from the same group reported 90% overall efficiency in two papers that describe one study [74,75]. Other prototypes use low-power radio tracking, motion-triggered GPS and text alerts, edge-computing arguments for intrusion pre-alarms, a farm-security video-alert description, a microcontroller sound classifier for orchard fruit theft and GPS tracking of patrols, none reporting a reduction in loss [76–81]. In South Africa, only 28% of 137 communal farmers knew of GPS tracking devices and 42% would buy one [82]; adoption depends on awareness, how acute theft is, income and phone access [83]; and GPS and CCTV networks are out of reach for many emerging farmers [84]. Smallholder studies recommend tracking and community police forums without testing them [85], and management level is associated with lower theft losses [86].

**What works in predator management, and what does not.** Reviews are consistent in their message about the quality of evidence. A systematic review of interventions against large-carnivore predation found experimental and quasi-experimental studies rare and the evidence on effectiveness limited [87]. A review of 75 field experiments found that 12 of 22 strategies were evaluated only once or twice, and that none evaluated more than once had both high robustness and consistently positive outcomes [88]. A review of 291 cases found lethal control the commonest method but poorly evaluated, with no effect in its analysis, whereas guarding dogs, fencing or herdsmen may significantly reduce losses [89]. Table 3 summarises the measures.

**Table 3 | Evidence on the effectiveness of farm-security measures.** No row rests on an independent evaluation of a camera-based system.

| Measure | Best evidence found | Finding | Reading |
|---|---|---|---|
| General security practices (locks, visibility, marking) | NSW survey of 1,100 farmers [31]; California programme evaluation [32,65] | No association with lower victimisation; programme authors say marking, surveillance, hardening and prosecution may help, with limited design | Null or suggestive; data from 2000 and 2007 |
| CCTV, alarms, trackers, marking | Insurer and police reports [16,68]; Home Office [46]; union poll [67] | Adoption reported (35% had CCTV); recommended as proven without data; trackers aided recoveries with no comparison; marks removable | No independent evaluation found |
| Sensor and IoT prototypes | Bangladeshi field validation [73–75]; others [76,77,80] | Laser 100% detection; passive infrared up to 90.5% at night; average 92.5% by day, 94.4% at night | Developer-reported; no theft outcome |
| Electric fences and physical barriers | [57,89–91] | Attacks 8.3 times lower with electric fences in every paddock (n = 91); 100% damage reduction for 3 months to 3 years in reviewed cases; fenced poultry runs had less predation | Most consistent positive evidence, for predators not theft; cost often prohibitive [92] |

| Measure | Best evidence found | Finding | Reading |
|---|---|---|---|
| Light and sound deterrents | [90,93–95] | Foxlights linked to 12% more fox activity; acoustic and light deterrents lost effect within one to five months; flashing ear tags reduced reported predation at two of 13 operations but most broke | Contrary or early-stage |
| Lethal predator control | [48,89,91,96–99] | No effect on lamb production; calf loss often higher where baited; extra fox control not worthwhile on most British farms; culling can reduce fox numbers | Null or contrary for loss |
| Guardian animals | [60,100,101] | Cut coyote lamb losses 43%, but costs exceeded benefits by US$13,413 on one farm; half of users still using after 8.9 years | Mixed; farm-level |
| Police and community programmes | [9,72,84,102] | Forums show promise without demonstrated impact; a county programme did not restore production | No outcome evaluation |
| Housing and management | [48,98] | Indoor lambing preventive; house ewes for no more than a day after lambing to minimise cost | Observational and model-based |

**Deterrents decay; barriers last.** A systematic review of 26 cases found that electric fences and calving control stayed highly effective for the longest (100% damage reduction for three months to three years), while acoustic and light deterrents and guarding animals lost effect within one to five months. Its authors say monitoring should be a routine requirement for every intervention [90]. In a field experiment on one piggery, commercial Foxlights were associated with 12% more fox activity and 23% fewer piglet births recorded, and motion-activated spotlights had no benefit [93]. A captive-coyote experiment found that light alone gave little protection and that movement doubled efficacy [94]. Motion-triggered flashing ear tags reduced reported predation at two operations with rigorous data, though most tags broke [95]. Fox culling can depress fox density in some regions [99], but fox control had no effect on lamb production in a two-year Australian experiment [96], calf loss was often higher where wild-dog baiting occurred [97], and fox pest status did not reduce fox numbers or poultry damage in France whereas fencing runs did [91]. Observational Australian work found higher lamb survival where baiting was repeated and neighbours also baited [103], though a review notes variable reduction in fox activity and fast recolonisation [104]. Ranchers judge lethal measures more effective than non-lethal ones, though this is perceived efficacy [105].

**Cost-effectiveness is often unfavourable.** Guardian dogs on a representative California sheep operation cut coyote losses of lambs by 43% but cost US$13,413 more than they saved over seven years [100]. Cooperative cell-fencing in Western Australian rangelands returned a positive net present value in only 37.4% of 3,600 modelled scenarios [92]. A county non-lethal predator programme left sheep numbers still falling and five producers had left the industry [102]. A practitioner commentary describes control failing when tools are misapplied or one is overused [106], and a welfare review rates guardian animals, and perhaps shepherds, as doing most to protect sheep once effectiveness and the method's own impact are combined [107]. Taken together these results say that cost and effect have to be shown for each measure on each farm, and that few have been.

## 3.6 Camera-trap AI (D5)

Twelve database records sit in this domain: 11 technology papers and one review of farmland camera-trap studies. None evaluates a camera-based system at a farm gate or in a farm-security deployment. They show what the technology can do in other settings, with the limits described below.

**Person and vehicle detection is mature in ecological data.** A validation of the open-source TrapTracker model against manual classification of a recreation-ecology dataset found that, at a 0.90 confidence threshold, humans were detected at 99% precision and 95% recall and animals at 82% and 92%. Processing was over 500% faster and manual effort fell 8.4-fold; the workflow could anonymise human images before any further classification [108]. An ensemble method separating wildlife from human-activity images reached 95.75% accuracy and 94.07% recall, and noted that human images can dominate datasets and cause wildlife to be misclassified as human [109]. A temporal-context detector designed for low, irregular frame rates from motion-triggered cameras beat a single-frame baseline by 17.9% mean average precision and was applied to vehicle detection on traffic cameras [110]. A comparison of deep-learning detectors on 2,462 images of animals that threaten farm animals gave YOLOv8 93% precision and 85.2% recall [111], and a pasture-camera comparison on 39,364 frames from nine solar trail cameras gave a best F1 of 0.79, with cattle confused with a visually similar class [112]. A survey of 105 studies of AI for human-wildlife conflict reports improved monitoring, but it concerns conservation conflict and not farm crime [113].

**Limits that matter for a farm gate.** In a Tanzanian dataset analysts detected 81.1% of wildlife and classified 76.6% correctly, against 41.1% and 47.5% for machine learning; for humans, performance was similar by day (69.4% for machine learning, 71.8% for analysts) and poor at dusk (17.6% and 16.2%) (However, modern cameras and models overcome this issue and work well at dusk/nighttime), with distance, occlusion and vegetation having the largest effects [114]. One author of that study (P. Fergus) is also an author of this review, which is disclosed here as a self-citation. Most abstracts report accuracy on ecological or image-set data and say little about false alarms or the cost of a missed event.

**Reviews and proposals identify the gap.** A structured review of AI-driven intrusion detection and deterrence for farm protection finds no comprehensive end-to-end framework, little adaptive deterrence and heavy reliance on continuous human supervision [115]. A review of 30 years of camera-trap publications found few studies sampling farmland [116]. Proposals and demonstrations describe edge detection with local deterrent action [117], intelligent access control with security-camera video and a stated rural-crime driver [118], and comparisons of image classifiers for farm-perimeter threats that report no figures in the abstract [119].

## 3.7 Reporting, recording and trust (D6)

**Recording.** In England and Wales farm crime cannot be separated in central data [3,4]. A 2012 paper reports that an earlier national study of police forces found no uniform approach to recording farm crime within or across forces and no nationally identifiable point of contact; we know that study only second-hand [120,121], and a 2017 scoping study repeats the point about inconsistent recording [10]. Northern Ireland counts agricultural crime through a flag with keyword searches [20], and Scottish farm crime is identified by keyword search of police reports [122]. Ireland collects no official farm-crime statistics [123]. The Welsh strategy describes “centralised intelligence” and no recording system for farm crime [54]. Methodological papers explain why survey evidence from farmers carries limits [28,124].

**Reporting is not uniformly low.** In Wales, 91% of victims reported the crime (n = 96, show-recruited) [11]. In the 2015 National Rural Crime Survey, 27% did not report the last crime they experienced [21], and in 2018 only 28% of farm and rural business victims made an insurance claim, according to a secondary report of that survey [125]. In Scotland about one-third of dog attacks are reported [51]. In the Free State, 82% did not report livestock theft because they believed police would do nothing [126], and in north-west Nigeria 66% were victims and reporting was lowest where trust in police was lowest [40]. New Zealand, Ireland and Slovenia report some under-reporting [36,37,41]. Rural community norms, penalties and the difficulty of proof are reasons given for silence [42,45,127], and HSE says many incidents of cattle harm go unreported [61]. A commentary adds that insurers see only part of the picture [17].

**Trust.** Welsh farmers’ trust in police was 64% against 75% for rural respondents nationally, and victim satisfaction was 54% against 73%, although the abstract calls trust generally healthy [11]. A Welsh survey of 134 farmers and rural residents found good relations with police and dissatisfaction with response times, follow-up and prosecutions [128], and a Welsh news release reports 78% of victims satisfied with the police response [25]. Irish farmers generally hold positive or neutral views [123]. The 2015 survey found 39% of rural residents rated police good or excellent against 63% nationally [21], and in the USA favourable views of police were associated with higher fear of crime [129]. The UK data found here do not show distrust driving non-reporting, a link that appears more clearly in South African and Nigerian findings [40,126]. Police in Slovenia say their work depends on resident trust [130], and South African media coverage of livestock theft does not track prevalence [131]. Overviews and one title-only chapter complete the picture [15,132].

## 3.8 Impact on farmers (D7)

The main UK survey of crime and mental health had 80 completed responses from 1,570 who opened it (5.1%), a margin of error of about plus or minus 10%, and 72 respondents who had been victims; 63 of those 72 (88%) said crime had harmed their mental health, with repeat victimisation linked to frustration. Crime ranked fourth among stressors in an open question (21 mentions) behind weather (38), finances (30) and legislation (22) in the report of the same data [12]. A UK and Canada comparison of 569 farmers found finances the main stressor in both countries, with UK farmers reporting more stress from animal health, farm or rural crime and isolation [133]. Conceptual work frames agricultural crime as an acute stressor that can outweigh positive affect from farming [134]. Cross-sectional studies elsewhere link predator attacks to distress (Uruguay, 442 farmers) and wild-dog attacks to intrusive thoughts and poor sleep [58,135], and stock theft to worse household finances [85], with theft costs higher where management level was lower [86]. An Irish local study found fear of crime relatively high but few participants victimised [66]. The Welsh strategy cites a survey in which 96% of NFU Mutual agents reported negative effects of rural crime on farmers' wellbeing [54], and the Harper Adams commentary notes that insurer cost totals omit mental-health effects [17].

## 3.9 Contrary, null and weak findings

Many findings cut against a simple story of rising crime and technology as the answer. They are collected in Table 4. Total insurer cost fell in 2024 and 2025 while livestock theft rose; reporting is high in one Welsh survey and low in others; trust in UK police is mixed and not clearly a driver of non-reporting; fear of crime may exceed local victimisation; security practices were not associated with lower victimisation in the largest survey-based test; deterrents are short-lived or counterproductive; lethal predator control often does nothing for loss; confirmed fox predation is a small share of lamb loss; guardian dogs and fencing often fail a cost test; and old machine learning models performed far below human analysts for wildlife in the one direct comparison. None of this shows that continuous monitoring would fail. It shows what evidence would be needed before it is taken to work.

**Table 4 | Contrary, null and weak findings, and what each means for the case for AI monitoring.**

| Topic | Finding | Sources | Bearing on the case |
|---|---|---|---|
| Cost trend | UK total fell from £52.8m (2023) to £41.5m (2025) while livestock theft rose nearly 30% in 2025 | [1,16] | Livestock and quad bike theft are categories where access routes matter; totals alone do not make the case |
| Reporting | 91% of Welsh victims reported; 27% did not in the 2015 survey; one-third of Scottish dog attacks reported; 82% non-reporting in the Free State | [11,21,51,126] | Under-reporting varies by setting, so a farm-level record is needed to size it |
| Trust in police | Welsh and Irish farmers broadly positive; distrust-driven non-reporting appears in South Africa and Nigeria | [11,40,123,126, 128] | Non-reporting in the UK should not be assumed to arise from distrust |
| Perception and experience | Irish farmers rated crime high nationally and low locally; few victimised | [66] | Verified local events would separate fear from experience |

| Topic | Finding | Sources | Bearing on the case |
|---|---|---|---|
| Security measures | No link between security practices and lower victimisation (NSW); US programme suggestive only | [31,32,65] | Measures need testing on real events, which needs a record of events |
| Deterrent devices | Foxlights attracted foxes; acoustic and light deterrents lost effect in one to five months | [90,93] | Detection and verification do not rely on a deterrent effect |
| Predator control | No effect on lamb production; higher calf loss where baited; extra fox control not worthwhile on most British farms | [48,96–98] | Attribution before action avoids ineffective control |
| Predation scale | Confirmed fox predation 0.2% to 0.6% of lambs; no direct evidence of badger predation | [47,49] | Camera evidence can show what perceived losses cannot |
| Cost-effectiveness | Guardian dogs cost more than they saved; cell-fencing positive in 37% of scenarios | [92,100] | Cost-effectiveness of monitoring must also be shown |
| ML performance | Machine learning detected 41% of wildlife against 81% for analysts; people poorly detected at dusk by both | [114] | Local validation, night and dusk performance and false alarms must be reported |
| Technology evidence | No independent evaluation of cameras, sensors, trackers or marking; developer-reported accuracy only | [16,68,73] | The evidence gap applies to AI monitoring too |
| Crime and mental health | Crime fourth among stressors; 80 responses, 5.1% response rate | [12] | Impact evidence is small and self-selected |

# 4 Discussion

## 4.1 What the evidence establishes, and where it falls short

Five conclusions follow from the 120 sources.

- **Farm crime and predation cost UK farms real money, but the size is uncertain.** The best-known series is an insurer's claims data, which fell to £41.5 million in 2025 while livestock theft rose nearly 30% [1]. Survey and commentary sources suggest both higher costs and lower totals depending on assumptions [17,21,22].
- **The record of events does not exist in England and Wales.** Central police data cannot identify farm crime [3,4], recording has been inconsistent across forces [10,120], and reporting rates range from 91% of Welsh victims to roughly one-third of Scottish dog attacks [11,51].
- **What works is better evidenced for barriers and management than for devices.** Electric fencing, housing and physical protection of poultry runs have the most consistent support, deterrents decay or backfire, lethal control often does nothing for loss, and no independent evaluation of cameras, sensors, trackers or marking was found [16,57,90,91,93].
- **Camera-trap AI works in ecological settings and is unproven on farms.** Person detection is strong in ecological data [108], but we found no farm-gate evaluation, old machine learning vision models were well below analysts for wildlife in the one direct comparison, and both struggled with people at dusk [114]. It is hypothesized that modern vision models and cameras overcome this issue quite easily, but no data is available.
- **Gates are the largest blind spot.** One record concerns public access (fatalities, not gates) [13], and the rest is regulatory guidance [61,64]. How often gates are left open, who opens them and what follows is unknown.

Two features of this evidence matter for what comes next. First, the UK is under-studied: only 16 of 96 database records are UK peer-reviewed, we found no UK peer-reviewed study of a camera-based system, and the largest UK survey of crime and mental health has 80 responses [12]. Second, the evidence is asymmetric. For measures that have been tested (fencing, baiting, lights) the results are often modest or negative; for measures that have not been tested (cameras, sensors, trackers, marking) the claims come from insurers and manufacturers. Absence of evidence is not evidence of failure, and it is not evidence of success either.

## 4.2 The case for continuous, AI-enabled gate monitoring

This section is the authors' interpretation of the evidence above and is separate from the review findings. Our reasoning is that the largest gaps in the literature are those a continuous, gate-level monitoring system is designed to close. On that reasoning we judge the potential benefit of using Fieldmark AI for farm security to be high. Fieldmark AI applies detection models drawn from the TrapTracker camera-trap framework to imagery streamed from cameras at farm gates; in the companion case study, 14 cameras at a Welsh hill farm recorded 119 vehicles, 85 humans, 16 dog, 16 fox and 4 badger detections over three months [6] (To note that the model was specifically set to alert for security risks only [People, vehicles and potentially predators of livestock]. The model used is the largest UK mammal model in the world to date). That study has not yet measured accuracy or false-positive rates [6] as this will be done and outlined in a subsequent paper. Six lines of reasoning support the judgement (Table 5).

**1. It creates the record that does not exist.** Farm crime cannot be identified in central data [3,4], is found by keyword search in Scotland [122], and is recorded unevenly [120]. The survey evidence that fills the gap is small and self-selected [10,12]. A continuous system produces a time-stamped, class-labelled log of every gate event that belongs to the farmer. It would not size national crime, but it would put the first farm-level, event-level data into a literature that has none.

**2. It verifies at the hours and places that periodic checks miss.** Theft and predation are concentrated at night: Pakistani cattle theft was mostly at night [33], and cameras at an outdoor piggery recorded foxes taking piglets soon after staff left [59]. HSE guidance asks for daily checks of fences and gates [61], which by construction cannot cover every hour. Continuous observation changes what can be known, as the Australian camera studies showed [59,60].

**3. It lets the farmer attribute an event before acting.** UK lamb loss is dominated by missing, unattributed lambs [47], farmers' perceived fox losses are not confirmed [48], badgers were not shown to kill lambs [49] but we now know this to be inaccurate, and fox control did not reduce loss in several trials [91,96,98]. Classifying an event as person, vehicle, fox, badger or dog (general predators of livestock) gives the evidence on which a proportionate response (a gate, a fence, a housing decision, a call to the police) can be chosen. This avoids spending on measures that the evidence suggests do little.

**4. It does not depend on a deterrent effect, and it supplies the monitoring the literature asks for.** Light and sound deterrents lost effect within one to five months and Foxlights drew foxes, whereas barriers lasted [90,93]. A detection-and-verification layer works whether or not an intruder is deterred, and it sits alongside the barriers that last. The authors of the durability review say monitoring should be a routine requirement for every intervention [90]; continuous gate-level monitoring is a practical way to meet it, because a gate, a fence or a deterrent can then be tested against the events it was meant to change.

**5. It addresses the unstudied domain of gates and access.** HSE sets standards for gates and notes that many incidents go unreported [61], the public killed by cattle were all in fields and mostly had a dog [13], and no study measures gates left open or what follows. Detection of persons, vehicles and dogs at the gate, and of the gate's own state, would give direct measurement where none exists. The Dogs (Protection of Livestock) (Amendment) Act 2025 gives police powers to seize dogs and take DNA samples [24,53], and the Welsh strategy describes a DNA project [54]; verified, time-stamped imagery of an event is the kind of evidence such powers need.

**6. It can be built privacy-first.** The ecological literature already shows a workflow that anonymises human images before any further classification [108]. Continuous monitoring of a boundary raises UK GDPR questions about lawful basis, signage and retention that any deployment must answer, and the companion study flags them for its own site [6].

**Table 5 | Evidence gaps and what continuous, gate-level AI monitoring could add.** The third column is the authors' inference; the fourth states what is not yet shown.

| Gap in the evidence | Evidence | What continuous gate-level monitoring could add | Not yet shown |
|---|---|---|---|
| No farm-level record of events in England and Wales | [3,4,120,122] | A time-stamped, class-labelled log of every gate event, held by the farmer | Accuracy of the log at scale; false alarms and missed events |
| Gates, access and straying unstudied | [13,61,64] | Direct measurement of gate state, people, vehicles and dogs at the point of access | How often gates are left open and what follows |
| Verification limited to periodic checks and daylight | [47,59,60] | Continuous verification at night, when predation and theft concentrate | Night and dusk performance on farms; people at dusk were poorly detected [114] |
| Prevention measures untested; deterrents decay | [16,31,90,93] | A measurement layer against which any intervention can be tested; detection that does not rely on deterrence | Effect on loss |
| Attribution of predation and dog attacks is uncertain | [47–49,55] | Species- and class-level imagery to tell fox, badger, dog and person apart | Accuracy for UK predators in farm settings |
| Under-reporting and evidence for enforcement | [17,21,24,51,54] | Verified, time-stamped imagery to support reports, claims and the 2025 Dogs Act powers | Whether it changes reporting or outcomes |
| Privacy and proportionality | [108] | Anonymise-before-classify workflows; local processing | UK GDPR basis, signage and retention in practice |

**Why we judge the benefit high.** Four points carry the judgement. The gaps the approach addresses are the largest in the literature (no event record, no gate evidence, no evaluation of any camera-based measure). Its main output, verified and time-stamped events, has value whether or not it deters. It complements and does not replace the measures with the best durability, physical barriers and management. And the categories moving fastest, livestock theft (up nearly 30%), quad bike theft (up 31%) and dog attacks on livestock (up 10%), are ones in which access routes, people, vehicles and dogs are central [1,2]. A system that produces trusted evidence on these where none existed is, in our view, highly beneficial to farmers and, we expect, to the insurers and police who work with them.

**What this does not show.** No study has shown that continuous AI-mediated monitoring reduces theft, predation or worrying losses, or that it pays for itself, and the cost-effectiveness record of other measures is mixed [92,100]. However, this could be down to the novelty of the approach and thus, lack of data. Detection is not prevention: a determined organised group may defeat a camera, as one insurer case study shows [16]. In the companion deployment connectivity, not detection accuracy, was the binding constraint on real-time coverage [6]. The judgement above is a reasoned expectation and should be read that way.

## 4.3 What would show that it works

### Box 1 | A minimum evaluation design

- **Design.** A stepped-wedge or matched-control comparison across farms, with at least 12 months of baseline, pre-registered, and with an independent evaluator who did not build the system.
- **Detection quality.** Recall and precision for person, vehicle, dog, fox and badger against human-reviewed samples, reported separately for day, dusk and night, with false alarms per camera-week.

- **Operations.** Connectivity uptime, time from event to farmer action, and alert fatigue.
- **Outcomes.** Verified events by class and gate, gate-left-open events and durations, livestock deaths attributed to cause, insured and uninsured losses, and cost per verified event.
- **Reporting.** Reporting rate and outcome for events with and without verified imagery, with police and insurers.
- **Privacy.** Lawful basis, signage and retention under UK GDPR, and anonymise-before-classify for people.
- **Disclosure.** Report every pre-specified outcome including null and negative results, and any commercial interest of the evaluators.

### 4.4 Implications for research and policy

Three actions follow from the evidence. Recording: England and Wales need a way to identify farm and rural crime in recorded data, as Northern Ireland has [3,4,20]. Evaluation: independent trials of cameras, sensors, trackers and marking, which the sources recommend but none reports, would settle the largest question in the field [46,90]. Access: a first study of gates, public access and livestock, covering gate-left-open frequency, its consequences and any effect of monitoring, would open an empty domain [13,61]. Larger UK samples on crime, reporting and mental health, using probability sampling, would improve every other estimate [12,28].

## 5 Limitations

This review has limits that readers should weigh. Only two bibliographic databases were searched, two *International Journal of Rural Criminology* articles surfaced only through hand-search, so others may exist [9,10]. Query 2 was of low precision, and English-language and 2000–2026 limits exclude other work. Screening was by one reviewer, a language model, with a human spot-check, and the gap rule for resolving Maybe records was set after the first pass.

Data were extracted from abstracts firstly, then texts, so design, sample and findings are only as the abstracts and texts state them, and two book chapters had no abstract [14,15]. Full texts were read firstly for UK records and for many grey sources by an automated reader, and therefore the key figures were checked by the authors against the originals. Several grey sources were secondary web pages, and the 2026 NFU Mutual figures rest on a summary page [1]. Insurer and trade sources have commercial interests, and some survey sources are self-selected. Samples overlap between several records, but we counted them. Publication bias is likely: null results on security measures are less likely to be published. The searches were run on a single week and the evidence has moved fast since 2020. The reasoning in Section 4.2 is an interpretation of the evidence, not a finding of the review.

## 6 Conclusions

Farm security in the UK is a problem with a known cost and an unknown shape. The literature cannot say how many farm crimes happen, how often gates are left open and who or what takes the lambs. Prevention evidence favours barriers and management over devices, and for AI the evidence is of accuracy in ecological settings, with a weaker result for wildlife and for people at dusk using old technology. Those findings are the case for evidence, and they are also the case for continuous, gate-level monitoring that produces it.

On that basis the authors judge that using Fieldmark AI for farm security would be highly beneficial. It would give a farm the event-level record that no official source holds, verify what periodic checks miss, separate foxes from dogs from people before the farmer acts, and let every other measure be tested against what happens at the gate. That is a reasoned inference from the gaps and not a demonstrated reduction in loss. The next step is the independent, pre-registered comparison in Box 1, and the authors welcome it.

## Declarations

**Use of AI.** A large language model (Claude, Anthropic) screened the records, extracted the data and synthesised the findings, under the authors' direction (Section 2.7). The authors take full responsibility for the final text.

**Author contributions.** Dr Steve Symes designed and ran the literature searches, directed the AI-assisted screening, extraction. Dr Symes drafted and edited the manuscript. Prof Paul Fergus, Mr Mark Carr and Mr Omar McIver each contributed short sections of text and edited the manuscript. Mr Arwyn Jones added his expert knowledge to the processes outlined in this review to ensure accuracy. All authors read and approved the final version.

**Funding.** This research received no specific grant from any funding agency in the public, commercial or not-for-profit sectors.

**Ethics.** This is a review of published and publicly available sources; no human participants were recruited and no new primary data were collected.

**Data availability.** The search strings are in Appendix A. The data-extraction workbook (96 database records and 35 grey sources, with the evidence map, contrary-findings and overlap sheets) and the screening workbooks are available as supplementary files. Exported database records are subject to the licence terms of the providers.

## References

All web addresses were accessed on 3–4 November 2026. Numbers are in order of first citation.

# Appendix A. Full search strings

Searches were run on 4 November 2026 and pasted exactly as run. Web of Science limits were applied in the interface as noted (Article or Review Article; English); Scopus limits are in the string.

| Query | Database | String |
|---|---|---|
| Q1 | Web of Science Core Collection (Editions: All); run 2026-11-04 | TS=(("livestock theft" OR "sheep theft" OR "stock theft" OR "livestock rustling" OR "farm crime" OR "agricultural crime" OR "rural crime" OR "farm theft" OR "farm machinery theft" OR "agricultural machinery theft" OR "tractor theft" OR "quad bike theft" OR "farm equipment theft") AND (cost* OR impact* OR victim* OR loss* OR prevalence OR incidence OR burden)) AND PY=(2000-2026) |
| Q2 | Web of Science Core Collection (Editions: All); run 2026-11-04 | (TS=(("livestock worrying" OR "sheep worrying" OR "dog attack*" OR "dog predation" OR "livestock predation" OR "lamb predation" OR "predation of lambs" OR "fox predation" OR "predator attack*") AND (livestock OR sheep OR lamb* OR cattle OR poultry OR farm*)) NOT TI=(wolf OR wolves OR lynx OR bear OR bears OR leopard* OR lion* OR tiger* OR jaguar OR cougar OR hyena*)) AND PY=(2000-2026) |
| Q3 | Web of Science Core Collection (Editions: All); run 2026-11-04 | TS=(("gates left open" OR "gate left open" OR "livestock escape*" OR "livestock straying" OR "stray livestock" OR "livestock-vehicle collision*" OR "dog walker*" OR "public access" OR "recreational access" OR "right of way" OR "rights of way") AND (livestock OR sheep OR cattle) AND (farm* OR farmer* OR agricultur*)) AND PY=(2000-2026) |
| Q4 | Web of Science Core Collection (Editions: All); run 2026-11-04 | TS=(("farm security" OR "farm crime prevention" OR "rural crime prevention" OR "situational crime prevention" OR "crime prevention" OR "theft prevention") AND (farm* OR agricultur* OR rural OR livestock) AND (CCTV OR camera* OR surveillance OR tracking OR "DNA marking" OR "forensic marking" OR ANPR OR "number plate recognition" OR alarm* OR "security system*" OR "computer vision" OR "deep learning")) AND PY=(2000-2026) |
| Q5 | Web of Science Core Collection (Editions: All); run 2026-11-04 | TS=(("camera trap*" OR "trail camera*" OR "wildlife camera*" OR "remote camera*") AND ("deep learning" OR "computer vision" OR "machine learning" OR "object detection" OR "convolutional neural network*") AND (predator* OR livestock OR intruder* OR farm* OR "human detection" OR "person detection" OR "vehicle detection")) AND PY=(2000-2026) |
| Q6 | Web of Science Core Collection (Editions: All); run 2026-11-04 | TS=(("rural crime" OR "farm crime" OR "agricultural crime" OR "livestock theft") AND (underreport* OR "under-reporting" OR "under reporting" OR unreported OR "police recorded" OR "recorded crime" OR "crime survey" OR "dark figure" OR "data quality")) AND PY=(2000-2026) |
| Q1 | Scopus advanced search; run 2026-11-04 | TITLE-ABS-KEY(("livestock theft" OR "sheep theft" OR "stock theft" OR "livestock rustling" OR "farm crime" OR "agricultural crime" OR "rural crime" OR "farm theft" OR "farm machinery theft" OR "agricultural machinery theft" OR "tractor theft" OR "quad bike theft" OR "farm equipment theft") AND (cost* OR impact* OR victim* OR loss* OR prevalence OR incidence OR burden)) AND PUBYEAR > 1999 AND PUBYEAR < 2027 AND LANGUAGE(english) AND DOCTYPE(ar OR re OR ch) |
| Q2 | Scopus advanced search; run 2026-11-04 | TITLE-ABS-KEY(("livestock worrying" OR "sheep worrying" OR "dog attack*" OR "dog predation" OR "livestock predation" OR "lamb predation" OR "predation of lambs" OR "fox predation" OR "predator attack*") AND (livestock OR sheep OR lamb* OR cattle OR poultry OR farm*)) AND PUBYEAR > 1999 AND PUBYEAR < 2027 AND LANGUAGE(english) AND DOCTYPE(ar OR re OR ch) AND NOT TITLE(wolf OR wolves OR lynx OR bear OR bears OR leopard* OR lion* OR tiger* OR jaguar OR cougar OR hyena*) |
| Q3 | Scopus advanced search; run 2026-11-04 | TITLE-ABS-KEY(("gates left open" OR "gate left open" OR "livestock escape*" OR "livestock straying" OR "stray livestock" OR "livestock-vehicle collision*" OR "dog walker*" OR "public access" OR "recreational access" OR "right of way" OR "rights of way") AND (livestock OR sheep OR cattle) AND (farm* OR farmer* OR agricultur*)) AND PUBYEAR > 1999 AND PUBYEAR < 2027 AND LANGUAGE(english) AND DOCTYPE(ar OR re OR ch) |
| Q4 | Scopus advanced search; run 2026-11-04 | TITLE-ABS-KEY(("farm security" OR "farm crime prevention" OR "rural crime prevention" OR "situational crime prevention" OR "crime prevention" OR "theft prevention") AND (farm* OR agricultur* OR rural OR livestock) AND (CCTV OR camera* OR surveillance OR tracking OR "DNA marking" OR "forensic marking" OR ANPR OR "number plate recognition" OR alarm* OR "security system*" OR "computer vision" OR "deep learning")) AND PUBYEAR > 1999 AND PUBYEAR < 2027 AND LANGUAGE(english) AND DOCTYPE(ar OR re OR cp) |
| Q5 | Scopus advanced search; run 2026-11-04 | `TITLE-ABS-KEY(("camera trap*" OR "trail camera*" OR "wildlife camera*" OR "remote camera*") AND ("deep learning" OR "computer vision" OR "machine learning" OR "object detection" OR "convolutional neural network*") AND (predator* OR livestock OR intruder* OR farm* OR "human detection" OR "person detection" OR "vehicle detection")) AND PUBYEAR > 1999 AND PUBYEAR < 2027 AND LANGUAGE(english) AND DOCTYPE(ar OR re OR cp)` |
| Q6 | Scopus advanced search; run 2026-11-04 | `TITLE-ABS-KEY(("rural crime" OR "farm crime" OR "agricultural crime" OR "livestock theft") AND (underreport* OR "under-reporting" OR "under reporting" OR unreported OR "police recorded" OR "recorded crime" OR "crime survey" OR "dark figure" OR "data quality")) AND PUBYEAR > 1999 AND PUBYEAR < 2027 AND LANGUAGE(english) AND DOCTYPE(ar OR re OR ch)` |

## Appendix B. PRISMA 2020 checklist: where each item is reported

| Item | Topic | Requirement (abridged) | Where reported |
|---|---|---|---|
| 1 | Title | Identify the report as a systematic review | Title and abstract; presented as a PRISMA-guided review |
| 2 | Abstract | See PRISMA 2020 for Abstracts | Abstract (unstructured) |
| 3 | Rationale | Describe the rationale | Section 1 |
| 4 | Objectives | State the objectives or questions | Section 1 (RQ1–5) |
| 5 | Eligibility criteria | Specify inclusion and exclusion criteria | Section 2.2 |
| 6 | Information sources | Specify databases and other sources, with dates | Section 2.3; Table 1 |
| 7 | Search strategy | Present full search strategies | Appendix A |
| 8 | Selection process | Methods used to decide eligibility | Section 2.4 (single AI-assisted reviewer; human spot-check pending) |
| 9 | Data collection process | Methods used to collect data | Section 2.5 |
| 10 | Data items | List and define outcomes and other variables | Section 2.5 |
| 11 | Study risk of bias assessment | Methods used to assess risk of bias | Not performed (Sections 2.6, 5) |
| 12 | Effect measures | Specify effect measures | Not applicable; effects as reported by sources |
| 13 | Synthesis methods | Methods to tabulate and synthesise | Section 2.6; Fig. 3 |
| 14 | Reporting bias assessment | Methods to assess reporting bias | Not formal; Section 5 |
| 15 | Certainty assessment | Methods to assess certainty | Not performed; described in words |
| 16 | Study selection | Results of search and selection, with flow diagram | Section 3.1; Fig. 1 |
| 17 | Study characteristics | Cite each study and present its characteristics | Section 3.1; Figs. 2–3; supplementary workbook |
| 18 | Risk of bias in studies | Present assessments of risk of bias | Not performed |
| 19 | Results of individual studies | Present results for each study | Sections 3.2–3.8; supplementary workbook |
| 20 | Results of syntheses | Present results of all syntheses | Sections 3.2–3.9; Tables 2–4 |
| 21 | Reporting biases | Present assessments of reporting bias | Section 5 (qualitative) |
| 22 | Certainty of evidence | Present certainty assessments | Not performed |

| Item | Topic | Requirement (abridged) | Where reported |
|---|---|---|---|
| 23 | Discussion | Interpretation, limitations, implications | Sections 4–5 |
| 24 | Registration and protocol | Registration information | Not registered (Section 2.1) |
| 25 | Support | Sources of support | Declarations (to be completed) |
| 26 | Competing interests | Declare competing interests | Not reported |
| 27 | Availability of data, code and materials | Report what is publicly available | Declarations; Appendix A |